%% file: main.tex
\documentclass{article}

\usepackage{microtype}
\usepackage{graphicx}
\usepackage{subfigure}
\usepackage{booktabs} 
\usepackage{amsmath}
\usepackage{amssymb}
\usepackage{lib}

\usepackage[colorlinks=true,linkcolor=red]{hyperref}

\usepackage[accepted]{icml2021}

\icmltitlerunning{PixelTransformer: Sample Conditioned Signal Generation}

\begin{document}

\twocolumn[
\icmltitle{PixelTransformer: Sample Conditioned Signal Generation}

\begin{icmlauthorlist}
\icmlauthor{Shubham Tulsiani}{fair}
\icmlauthor{Abhinav Gupta}{fair,cmu}
\end{icmlauthorlist}

\icmlaffiliation{fair}{Facebook AI Research}
\icmlaffiliation{cmu}{Carnegie Mellon University}

\icmlcorrespondingauthor{Shubham Tulsiani}{shubhtuls@fb.com}

\begin{center}
{ \small \tt \href{https://shubhtuls.github.io/PixelTransformer/}{https://shubhtuls.github.io/PixelTransformer/}}
\end{center}

\vskip 0.2in
]

\printAffiliationsAndNotice{}

\begin{abstract}
We propose a generative model that can infer a distribution for the underlying spatial signal conditioned on sparse samples \eg plausible images given a few observed pixels. In contrast to sequential autoregressive generative models, our model allows conditioning on arbitrary samples and can answer distributional queries for any location. We empirically validate our approach across three image datasets and show that we learn to generate diverse and meaningful samples, with the distribution variance reducing given more observed pixels. We also show that our approach is applicable beyond images and can allow generating other types of spatial outputs \eg polynomials, 3D shapes, and videos.
\end{abstract}

\addtolength{\textfloatsep}{-4mm}
\newcommand{\val}{\mathbf{v}}
\newcommand{\pos}{\mathbf{x}}
\newcommand{\posg}{g}
\newcommand{\R}{\mathbb{R}}

\input{intro}
\input{formulation}
\input{related}
\input{approach}
\input{results}
\input{nd}

\input{discussion}

\vspace{-1mm}
\noindent \textbf{Acknowledgements.} We would like to thank Deepak Pathak and the members of the CMU Visual Robot Learning lab for helpful discussions and feedback.

\bibliography{main}
\bibliographystyle{icml2021}

\newpage
\input{appendix}

\end{document}

%% file: intro.tex
\vspace{-1mm}
\section{Introduction}
\vspace{-1mm}
Imagine an artist with an empty canvas. She starts with a dab of sky blue paint at the top, and a splash of fresh green at the bottom. What is the painting going to depict? Perhaps an idyllic meadow, or trees in garden under a clear sky? But probably not a living room. It is quite remarkable that given only such sparse information about arbitrary locations, we can make guesses about the image in the artist's mind. 

The field of generative modeling of images, with the goal of learning the distribution of possible images, focuses on developing similar capabilities in machines. Most recent approaches can be classified as belonging to one of the two modeling frameworks. First, and more commonly used, is the latent variable modeling framework~\cite{vae,gan}. Here, the goal is to represent the possible images using a distribution over a bottleneck latent variable, samples from which can be decoded to obtain images. However, computing the exact probabilities for images is often intractable and it is not straightforward to condition inference on sparse observations \eg pixel values. As an alternative, a second class of autoregressive  approaches directly model the joint distribution over pixels. This can be easily cast as product of conditional distribution~\cite{pixelcnn,pixelrnn} which makes it tractable to compute. Conditional distributions are estimated by learning to predict new pixels from previously sampled/generated pixels. However, these approaches use fixed sequencing (mostly predicting pixels from top-left to bottom-right) and therefore the learned model can only take a fixed ordering between query and sampled pixels. This implies that these models cannot predict  whole images from a few random splashes -- similar to what we humans can do given a description of the artist's painting above.

In this work, our goal is to build  computational generative models that can achieve this -- given information about some \emph{random} pixels and their associated color values, we aim to predict a \emph{distribution} over images consistent with the evidence. To this end, we show that it suffices to learn a function that estimates the distribution of possible values at any query location conditioned on an arbitrary set of observed samples. We present an approach to learn this function in a self-supervised manner, and show that it can allow answering queries that previous sequential autoregressive models cannot \eg mean image given observed pixels, or computing image distribution given random observations. We also show that our proposed framework is generally applicable beyond images and can be learned to generate generic dense spatial signals given corresponding samples.

%% file: formulation.tex
\vspace{-1mm}
\section{Formulation}
\vspace{-1mm}
\seclabel{formulation}
Given the values of some (arbitrary) pixels, we aim to infer what images are likely conditioned on this observation.
More formally, for any pixel denoted by random variable $\pos$, let $\val_{\pos}$ denote the value for that pixel and let $S_0 \equiv \{\val_{\pos_k}\}_{k=1}^K$ correspond to a set of such sampled values. We are then interested in modeling $p(I | S_0)$ \ie the conditional distribution over images $I$ given a set of sample pixel values $S_0$.

\noindent \textbf{From Image to Pixel Value Distribution.}
We first note that an image is simply a collection of values of pixels in a discrete grid. Assuming an image has $N$ pixels with locations denoted as $\{\posg_n\}_{n=1}^N$, our goal is therefore to model $p(I|S_0) \equiv p(\val_{\posg_1}, \val_{\posg_2}, \dots, \val_{\posg_N} | S_0)$. Instead of modeling this joint distribution directly, we observe that it can be further factorized as a product of conditional distributions using the chain rule:
\begin{gather*}
    p(\val_{\posg_1}, \val_{\posg_2}, \dots, \val_{\posg_N} | S_0) = \prod_n p(\val_{\posg_n} | S_0, \val_{\posg_1}, \dots, \val_{\posg_{n-1}})
\end{gather*}
Denoting by $S_n \equiv S_0 \cup \{\val_{\posg_j}\}_{j=1}^{n}$, we obtain:
\begin{gather}
\eqlabel{pis}
    p(I | S_0) = \prod_n p(\val_{\posg_n} | S_{n-1})
\end{gather}

\vspace{-2mm}
\noindent \textbf{Sample Conditioned Value Prediction.} The key observation from \eqref{pis} is that all the factors are in the form of $p(\val_\pos | S)$. That is, the only queries we need to answer are: `\emph{given some observed samples $S$, what is the distribution of possible values at location $\pos$}'?
To learn a sample conditioned generative model for images, we therefore propose to learn a function $f_{\theta}$ to infer $p(\val_\pos | S)$ for arbitrary inputs $\pos$ and $S$. Concretely, we formulate our task as that of learning a function $f_{\theta}(\pos, \{(\pos_k, \val_k)\})$ that can predict the value distribution at an arbitrary query location $\pos$ given a set of arbitrary sample (position, value) pairs $\{(\pos_k, \val_k)\}$.

\noindent In summary:
\vspace{-2mm}
\begin{itemize}
\vspace{-1mm}
    \item The task of inferring $p(I|S_0)$ can be reduced to queries of the form $p(\val_\pos | S)$.
    \vspace{-1mm}
\item We propose to learn a function $f_{\theta}(\pos, \{(\pos_k, \val_k)\})$ that can predict $p(\val_\pos | \{\val_{\pos_k}\})$ for arbitrary inputs.
\end{itemize}

\vspace{-2mm}
While we used images as a motivating example, our formulation is also applicable for modeling distributions of other dense spatially varying signals. For RGB images, $\pos \in \mathbb{R}^2, \val \in \mathbb{R}^3$, but other spatial signals \eg polynomials ($\pos \in \mathbb{R}^1, \val \in \mathbb{R}^1$), 3D shapes represented as Signed Distance Fields, ($\pos \in \mathbb{R}^3, \val \in \mathbb{R}^1$) or videos ($\pos \in \mathbb{R}^3, \val \in \mathbb{R}^3$) can also be handled by learning $f_{\theta}(\pos, \{(\pos_k, \val_k)\})$ of the corresponding form (see \secref{nd}).

%% file: related.tex
\vspace{-1mm}
\section{Related Work}
\vspace{-1mm}
\noindent \textbf{Autoregressive Generative Models.} Closely related to our work, autoregressive generative modeling approaches also factorize the joint distribution into per-location conditional distributions. Seminal works such as Wavenet~\cite{wavenet}, PixelRNN~\cite{pixelrnn} and PixelCNN~\cite{pixelcnn} showed that we can learn the distribution over the values of the `next' timestep/pixel given the values of the previous ones, and thereby learn a generative model for the corresponding domain (speech/images). Subsequent approaches have further improved over these works by modifying the parametrization~\cite{pixelcnnplus}, incorporating hierarchy~\cite{vqvae,vqvae2}, or  (similar to ours) foregoing convolutions in favor of alternate base architectures~\cite{chen2020generative,imTransformer} such as Transformers~\cite{transformer}.

While this line of work has led to impressive results, the core distribution modeled is that of the `next' value given `previous' values. More formally, while we aim to predict $p(\val_\pos | S)$ for \emph{arbitrary} $\pos, S$, the prior autoregressive generative models only infer this for cases where $S$ contains pixels in some sequential (\eg raster) order and $\pos$ is the immediate `next' position. Although using masked convolutions can allow handling many possible inference orders~\cite{jain2020uai}, the limited receptive field of convolutions still limits such orders to locally continuous sequences. Our work can therefore be viewed as a generalization of previous `sequential' autoregressive models in two ways: a) allowing  \emph{any} query position $\pos$, and b) handling arbitrary samples $S$ for conditioning. This allows  us to answer questions that prior autoregressive models cannot \eg `if the top-left pixel is blue, how likely is the bottom-right one to be green?', `what is the mean image given some observations?', or `given values of 10 specific pixels, sample likely images'.

\noindent \textbf{Implicit Neural Representations.} There has been a growing interest in learning neural networks to represent 3D textured scenes~\cite{deepvoxels}, radiance fields~\cite{nerf,martinbrualla2020nerfw,zhang2020nerf} or more generic spatial signals~\cite{siren,fourierfeat}. The overall approach across these methods is to represent the underlying signal by learning a function $g_{\phi}$ that maps query positions $\pos$ to corresponding values $\val$ (\eg pixel location to intensity). Our learned $f_{\theta}(\cdot, \{(\pos_k, \val_k)\})$ can similarly be thought of as mapping query positions to a corresponding value (distribution), while being conditioned on some sample values. A key difference however, is the ability to generalize --  the above mentioned approaches learn an independent network per instance \eg a separate $g_{\phi}$ is used to model each scene, therefore requiring from thousands to millions of samples to fit $g_{\phi}$ for a specific scene. In contrast, our approach uses a common $f_{\theta}$ across all instances and can therefore generalize to unseen ones given only a sparse set of samples. Although some recent approaches~\cite{disn,deepsdf,occnet} have shown similar ability to generalize and infer novel 3D shapes/scenes given input image(s), these cannot handle sparse input samples and do not allow inferring a distribution over the output space.

\noindent \textbf{Latent Variable based Generative Models.} Our approach, similar to sequential autoregressive models, factorizes the image distribution as products of per-pixel  distributions.  An alternate approach to generative modeling, however, is to transform a prior distribution over latent variables to the output distribution via a learned decoder. Several approaches allow learning such a decoder by leveraging diverse objectives \eg adversarial loss~\cite{gan}, variational bound on the log-likelihood~\cite{vae}, nearest neighbor matching~\cite{glo,imle}, or the  log-likelihood with a restricted decoder~\cite{rezende2015variational}. While all of these methods allow efficiently generating new samples from scratch (by randomly sampling in the latent space), it is not straightforward to condition this sampling given partial observations -- which is the goal of our work.

\noindent \textbf{Bayesian Optimization and Gaussian Processes.} As alluded to earlier, any spatial signal can be considered a function from positions to values. Our goal is then to infer a distribution over possible functions given a set of samples. This is in fact also a central problem tackled in bayesian optimization~\cite{brochu2010tutorial}, using techniques such as gaussian processes~\cite{rasmussen2003gaussian} to model the distribution over functions. While the goal of these approaches is similar to ours, the technique differs significantly. These classical methods assume a known prior over the space of functions and leverage it to obtain the posterior given some samples (we refer the reader to ~\cite{murphy2012machine} for an excellent overview). Such a prior over functions (that also supports tractable inference), however, is not easily available for complex signals such as images or 3D shapes -- although some weak priors~\cite{Ulyanov_2018_CVPR,Osher2017LowDM} do allow impressive image restoration, they do not enable generation given sparse samples. In contrast, our approach allows learning from data, and can be thought of as learning this prior as well as performing efficient inference via the learned model $f_{\theta}$.

%% file: approach.tex
\vspace{-1mm}
\section{Learning and Inference}
\vspace{-1mm}
Towards inferring the distribution of images given a set of observed samples, we presented a formulation in \secref{formulation} that reduced this task to that of learning a function to model $p(\val_\pos | \{\val_{\pos_k}\})$. We first describe in \secref{learning} how we parametrize this function and how one can learn it from raw  data. We then show in \secref{mean} and \secref{sampling} how this learned function can be used to query and draw samples from the conditional distribution over images $p(I | S_0)$. While we use images as the running example, we reiterate that the approach is more generally applicable (as we also empirically show in \secref{nd}).

\vspace{-1mm}
\subsection{Learning to Predict Value Distributions}
\vspace{-1mm}
\seclabel{learning}
We want to learn a function $f_{\theta}$ that can predict the probability distribution of possible values at any query location $\pos$ conditioned on a (arbitrary) set of positions with known values. More formally, we want $f_{\theta}(\pos, \{(\pos_k, \val_k)\})$ to approximate $p(\val_\pos | \{\val_{\pos_k}\})$.

\begin{figure}[t]
    \centering
    \includegraphics[width=.95\linewidth]{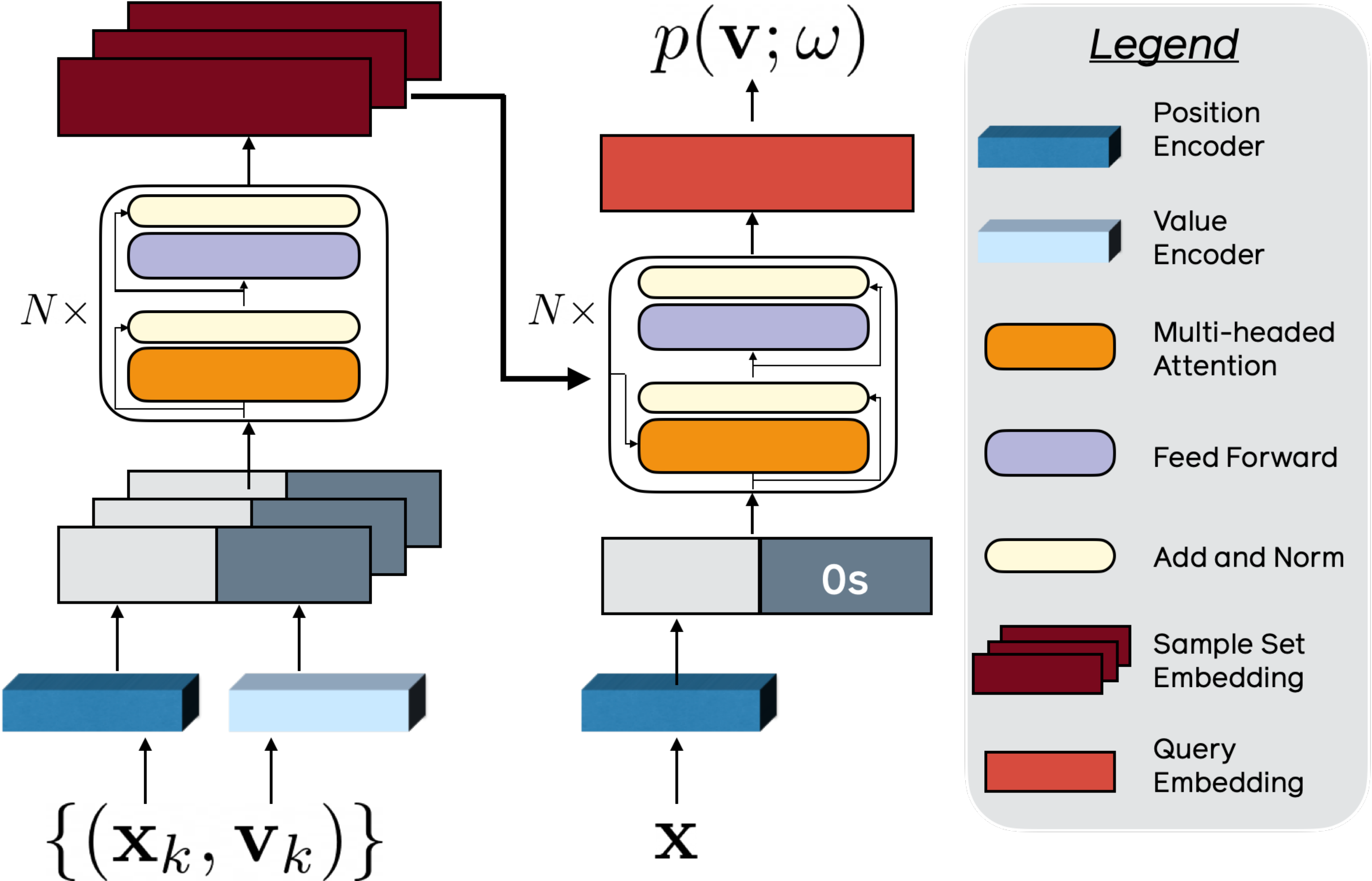}
    \vspace{1mm}
    \caption{\textbf{Prediction Model.} Given a set of (position, value) pairs $\{(\pos_k, \val_k)\}$, our model encodes them using a Transformer~\cite{transformer} encoder. A query position $\pos$ is then processed in context of this encoding and a value distribution is predicted (parametrized by $\omega$).}
    \figlabel{model}
\end{figure}

\noindent \textbf{Distribution Parametrization.} The output of $f_{\theta}$ is supposed to be a \emph{distribution} over possible values at location $\pos$ and not a single value estimate. How should we parametrize this distribution? Popular choices like gaussian parametrization may not capture the multimodal nature of the distribution \eg a pixel maybe black or white, but not gray. An alternate is to discretize the output space but this may require a large number of bins \eg $256^3$ for possible RGB values. Following  PixelCNN++~\cite{pixelcnnplus}, we opt for a hybrid approach -- we predict probabilities for the value belonging to one of $B$ discrete bins, while also predicting a continuous gaussian parametrization within each bin. This allows predicting multimodal distributions while enabling continuous outputs.

Concretely, we instantiate $B$ bins (roughly) uniformly spaced across the output space where for any bin $b$, its center corresponds to $c^b$. The output distribution is then parametrized as $\omega \equiv \{(q^b, \mu^b, \sigma^b)\}_{b=1}^B$. Here $q^b \in \R^1$ is the probability of assignment to bin $b$, $c^b + \mu^b$ is the mean of the corresponding gaussian distribution with uniform variance $\sigma^b \in \R^1$. Assuming the values $\val \in \R^d$, our network therefore outputs $\omega \in \R^{B \times (d + 2)}$. We note that this distribution is akin to a mixture-of-gaussians, and given a value $\val$, we can efficiently compute its likelihood $p(\val; \omega)$ under it (see appendix for details). We can also efficiently compute the expected value $\bar{\val}$ as:
\begin{gather}
\eqlabel{meanval}
    \bar{\val} \equiv \int p(\val; \omega) ~\val~ d\val = \sum_{b=1}^B q^b (\mu^b + c^b)
\end{gather}
\noindent \textbf{Model Architecture.}
Given a query position $\pos$, we want $f_{\theta}(\pos, \{(\pos_k, \val_k)\})$ to output a value distribution as parametrized above. There are two design considerations that such a predictor should respect: a) allow a variable number of input samples $\{(\pos_k, \val_k)\}$, and b) be permutation-invariant w.r.t. the samples. We leverage the Transformer~\cite{transformer} architecture as our backbone as it satisfies both these requirements. As depicted in \figref{model}, our model can be considered as having two stages: a) an encoder that, independent of the query $\pos$, processes the input samples $\{(\pos_k, \val_k)\}$ and computes a per-sample embedding, and b) a decoder that predicts the output distribution by processing the query $\pos$ in context of the encodings.

As shown in \figref{model}, we first independently embed each input sample $(\pos_k, \val_k)$ using position and value encoding modules respectively, while following the insight from \cite{fourierfeat} to use fourier features when embedding positions. These per-sample encodings are then processed by a sequence of multi-headed self-attention modules~\cite{transformer} to yield the encoded representations for the input samples. The query position $\pos$ is similarly embedded, and processed via  multi-headed attention modules in context of the sample embeddings. A linear decoder finally predicts $\omega \in \R^{B \times (d + 2)}$ to parametrize the output distribution.
\begin{figure}[t]
    \centering
    \includegraphics[width=.9\linewidth]{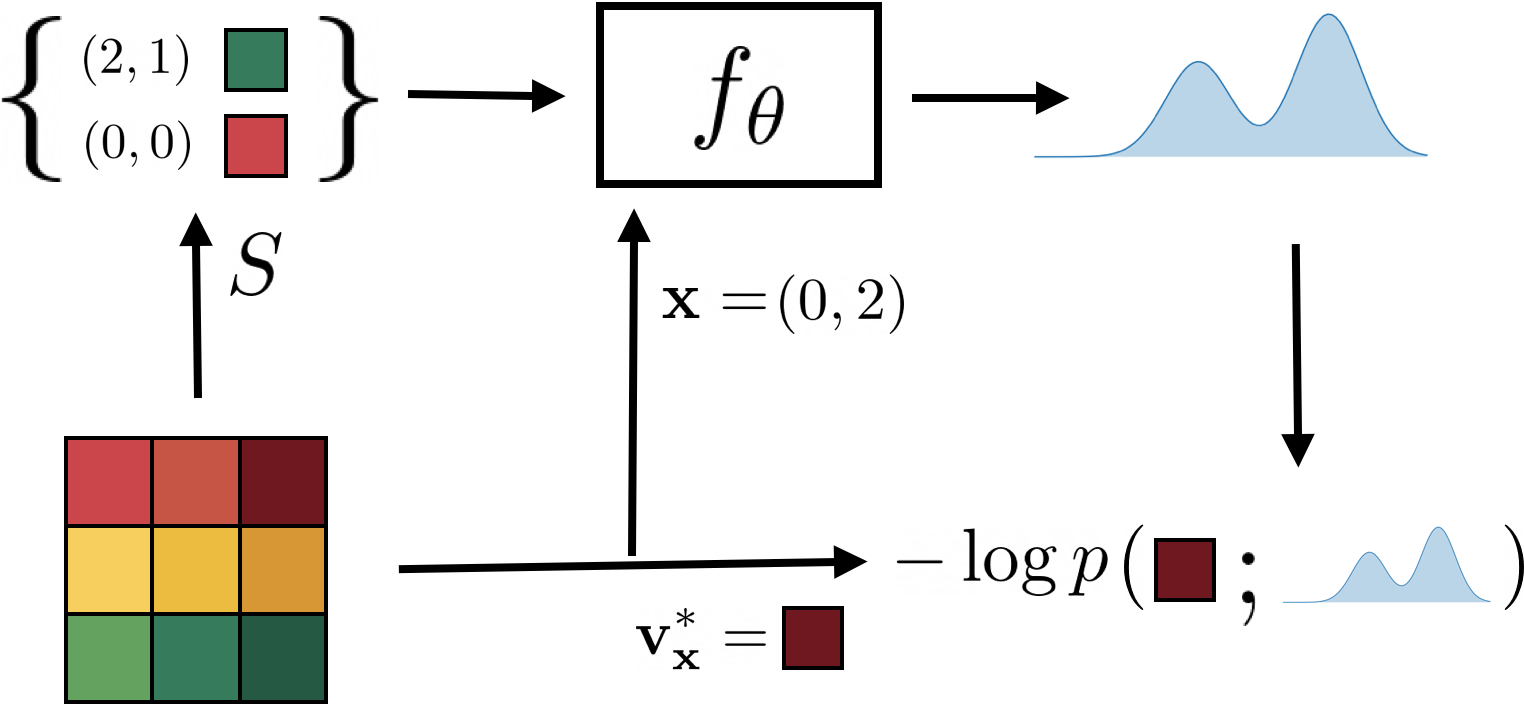}
    \vspace{1mm}
    \caption{\textbf{Training Overview.} Given an image, we randomly sample pixels to obtain the conditioning set $S$ as well as a query pixel $\pos$ with value $\val^*_{\pos}$. Our model predicts the conditional value distribution for this arbitrary query location and we use the negative log-likelihood for the true value as our learning objective.}
    \figlabel{training}
\end{figure}

\noindent \textbf{Training Objective.}
Recall that our model $f_{\theta}(\pos, \{(\pos_k, \val_k)\})$ aims to approximate $p(\val_\pos | \{\val_{\pos_k}\})$ for arbitrary query positions $\pos$ and sample sets $S \equiv \{\val_{\pos_k}\}$. Given a collection of training images, we can in fact generate training data for this model in a self-supervised manner. As illustrated in \figref{training}, we can simply sample arbitrary $\pos, S$ from any image, and maximize the log-likelihood of the true value $\val^{*}_{\pos}$ under the predicted distribution $p(\val_\pos | \{\val_{\pos_k}\})$.

While we described the processing for a single query position $\pos$, it is easy to parallelize inference and process a batch of queries $Q$ 
conditioned on the same input sample set $S$. In this case, we can consider the model as independently predicting $p(\val_\pos | \{\val_{\pos_k}\})$ for each $\pos \in Q$. Instead of using a single query $\pos$, we therefore use a batch of queries $Q$ and minimize the negative log-likelihood across them. More formally, given a dataset $D$ of images, we randomly sample an image $I$, and then choose arbitrary sample and query sets $S, Q$, and minimize the expected negative log-likelihood of the true values as our training objective:
\begin{equation}
\begin{split}
L = \mathop{\mathbb{E}}_{I \sim D} ~ \mathop{\mathbb{E}}_{S,Q \sim I} ~ \mathop{\mathbb{E}}_{\pos \sim Q} ~ - \log p(\val^*_\pos; \omega)
\\ \text{where},  \omega = f_{\theta} (\pos; \{(\pos_k, \val_k)\})
\end{split}
\end{equation}

\vspace{-1mm}
\subsection{Inferring Marginals and Mean}
\vspace{-1mm}
\seclabel{mean}
\secref{learning} introduced our approach to enable learning $f_{\theta}$ that can approximate $p(\val | S)$.  But given such a learned function, what can it enable us to do? One operation that we focus on later in \secref{sampling} is that of sampling images $I 
\sim p(I|S)$. However, there is another question of interest which is not possible to answer with the previous sequential autoregressive models~\cite{pixelcnn,wavenet}, but is efficiently computable using our model: `\emph{what is the expected image $\bar{I}$ given the samples $S$?}'.

We reiterate that an image can be considered as a collection of values of pixels located in a discrete grid $\{\posg_n\}_{n=1}^N$. Instead of asking what the expected image $\bar{I}$ is, we can first consider a simpler question -- what is the expected value ${\bar{\val}_{\posg_{n}}}$ for the pixel $\posg_n$ given $S$? By definition:
\begin{gather*}
    {\bar{\val}_{\posg_{n}}} = \int p(\val_{\posg_{n}} | S)~\val_{\posg_{n}}~d \val_{\posg_{n}}
\end{gather*}
As our learned model $f_{\theta}$ allows us to directly estimate the marginal distribution $p(\val_{\posg_{n}} | S)$, the above computation is extremely efficient to perform and can be done independently across all locations in the image grid $\{\posg_n\}_{n=1}^N$.
\begin{gather}
\eqlabel{mean}
    {\bar{\val}_{\posg_{n}}} = \int p(\val; \omega_n) ~\val~ d\val; ~~ \omega_n = f_{\theta}(\posg_n, \{(\pos_k, \val_k)\})
\end{gather}
Given the estimate of ${\bar{\val}_{\posg_{n}}}$, the mean image $\bar{I}$ is then just the image with each pixel assigned its mean value $\bar{\val}_{\posg_{n}}$ \ie $\bar{I} \equiv \{\bar{\val}_{\posg_{n}}\}_{n=1}^N$. The key difference compared to sequential autoregressive models~\cite{pixelcnn,wavenet} that enables our model to compute this mean image is that our model allows computing $p(\val_{\posg_{n}} | S)$ for \emph{any} location $\posg_{n}$, whereas approaches like ~\cite{pixelcnn,wavenet} can only do so for the `next' pixel.

\begin{figure*}
    \centering
    \includegraphics[width=.95\linewidth]{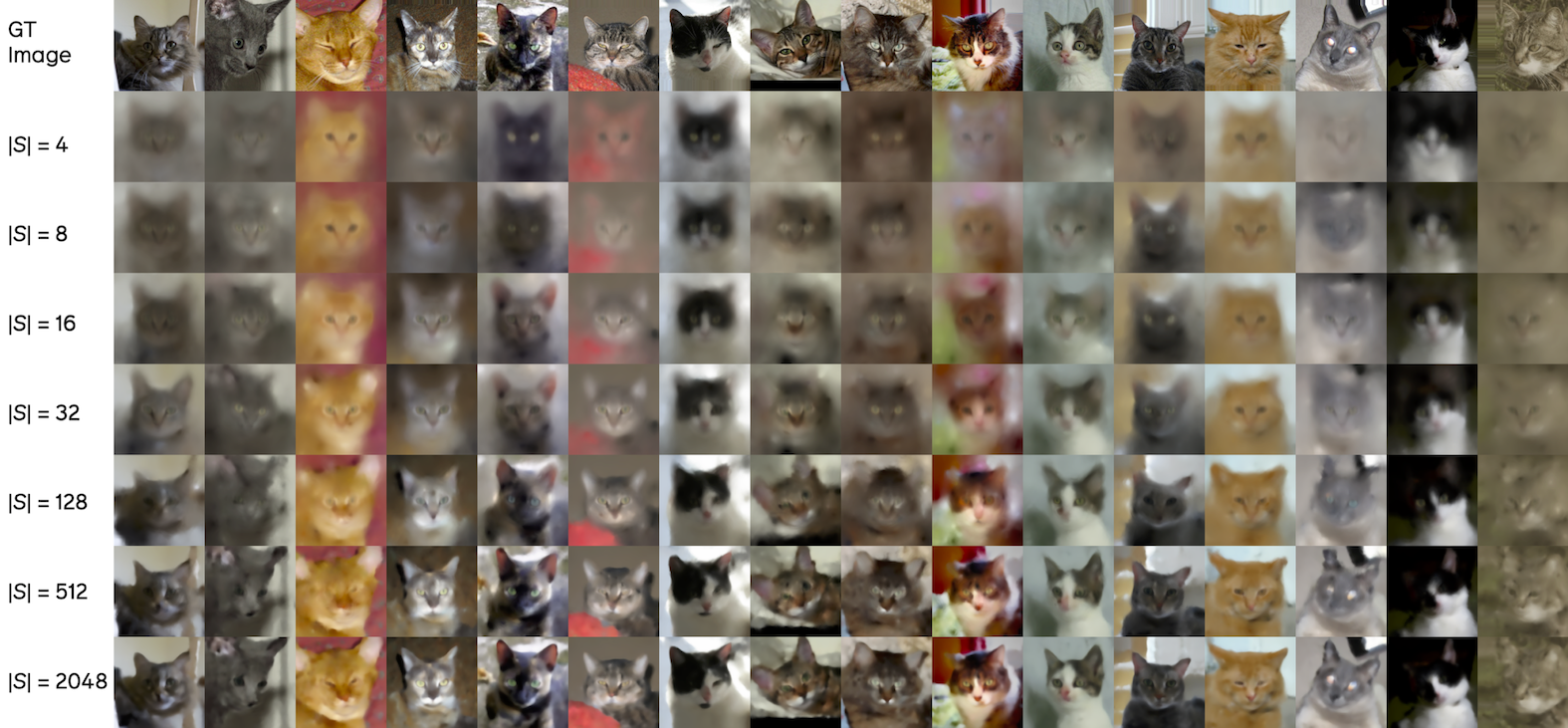}
    \vspace{1mm}
    \caption{\textbf{Inferred Mean Images.} We visualize the mean image predicted by our learned model on random instances of the Cat Faces dataset. Top row: ground-truth image. Rows 2-8: Predictions using increasing number of observed pixels $|S|$.}
    \figlabel{mean_cat}
\vspace{-4mm}
\end{figure*}

\vspace{-2mm}
\subsection{Autoregressive Conditional Sampling}
\vspace{-1mm}
\seclabel{sampling}
One of the driving motivations for our work was to be able to sample the various likely images conditioned on a sparse set of pixels with known values. That is, we want to be able to draw samples from $p(I|S_0)$. Equivalently, to sample an image from $p(I|S_0)$, we need to sample the values at each pixel $\{\val_{\posg_n}\}$ from $p(\val_{\posg_1}, \val_{\posg_2}, \dots, \val_{\posg_N} | S_0)$.

As we derived in \eqref{pis}, this distribution can be factored as a product of per-pixel conditional distributions. We can therefore sample from this distribution autoregressively -- sampling one pixel at a time, with subsequent pixels being informed by ones sampled prior. Concretely, we iteratively perform the following computation:
\begin{eqnarray}
\eqlabel{samplingdist}
    \omega_n = f_{\theta}(\posg_n, \{\pos_k, \val_k\} \cup \{\posg_j, \val'_j\}_{j=1}^{n-1})
\\ \val'_n \sim p(\val; \omega_n)
\eqlabel{samplingval}
\end{eqnarray}

Here, $\omega_n$ denotes the parameters for the predicted distribution for the pixel $\posg_n$. Note that this prediction takes into account not just the initial samples $S_0$, but also the subsequent $n-1$ samples (hence the difference from $\omega_n$ in \eqref{mean}).  $\val'_n$ represents a value then sampled for the pixel $\posg_n$ from the distribution parametrized by $\omega_n$.

\vspace{-1mm}
\noindent \textbf{Randomized Sampling Order.} While we sample the values one pixel at a time, the ordering of pixels $\posg_1, \dots, \posg_N$ need not correspond to anything specific \eg it is not necessary that $\posg_1$ should be the top-left pixel and $\posg_N$ be the bottom-right one. In fact, as our model $f_{\theta}$ is trained using arbitrary sets of samples $S$, using a structured sampling ordering \eg raster order would make the testing setup differ from training. Instead, for every sample $I \sim p(I|S)$ that we draw, we use a new \emph{random} order in which the pixels of the image grid are sampled.

\vspace{-1mm}
\noindent \textbf{Sidestepping Memory Bottlenecks.} As \eqref{samplingdist} indicates, the input to $f_{\theta}$ when sampling the $(n+1)^{th}$ pixel is a set of size $K + n$ -- the initial $K$ observations and the subsequent $n$ samples. Unfortunately, our model's memory requirement, due to the self-attention modules, grows cubically with this input size. This makes it infeasible to autoregressively sample a very large number of pixels. However, we empirically observe that given a sufficient number of (random) samples, subsequent pixel value distributions do not exhibit a high variance. We leverage this observation to design a hybrid sampling strategy. When generating an image with $N$ pixels, we sample the first $N'$  (typically 2048) autoregressively \ie following $\eqref{samplingdist}$ and $\eqref{samplingval}$. For the remaining $N-N'$ pixels, we simply use their mean value estimate conditioned on the initial and generated $K+N'$ samples (using \eqref{mean}). While this may lead to some loss in detail, we qualitatively show that the effects are not prohibitive and that the sample diversity is preserved.

%% file: results.tex
\vspace{-1mm}
\section{Experiments}
\vspace{-1mm}
To qualitatively and quantitatively demonstrate the efficacy of our approach, we consider the task of generating images given a set of pixels with known values. The goal of our experiments is twofold -- a) to validate that our predictions account for the observed pixels, and b) to show that the generated samples are diverse and plausible.

\vspace{-1mm}
\noindent \textbf{Datasets.} We examine our approach on three different image datasets -- CIFAR10~\cite{krizhevsky2009learning}, MNIST~\cite{mnist}, and the Cat Faces~\cite{catfaces} dataset while using the standard image splits. Note that we only require the images for training -- class or attribute labels are not leveraged for learning our models \ie even on CIFAR10, we learn a class-agnostic generative model.

\begin{figure*}
    \centering
    \includegraphics[width=.95\linewidth]{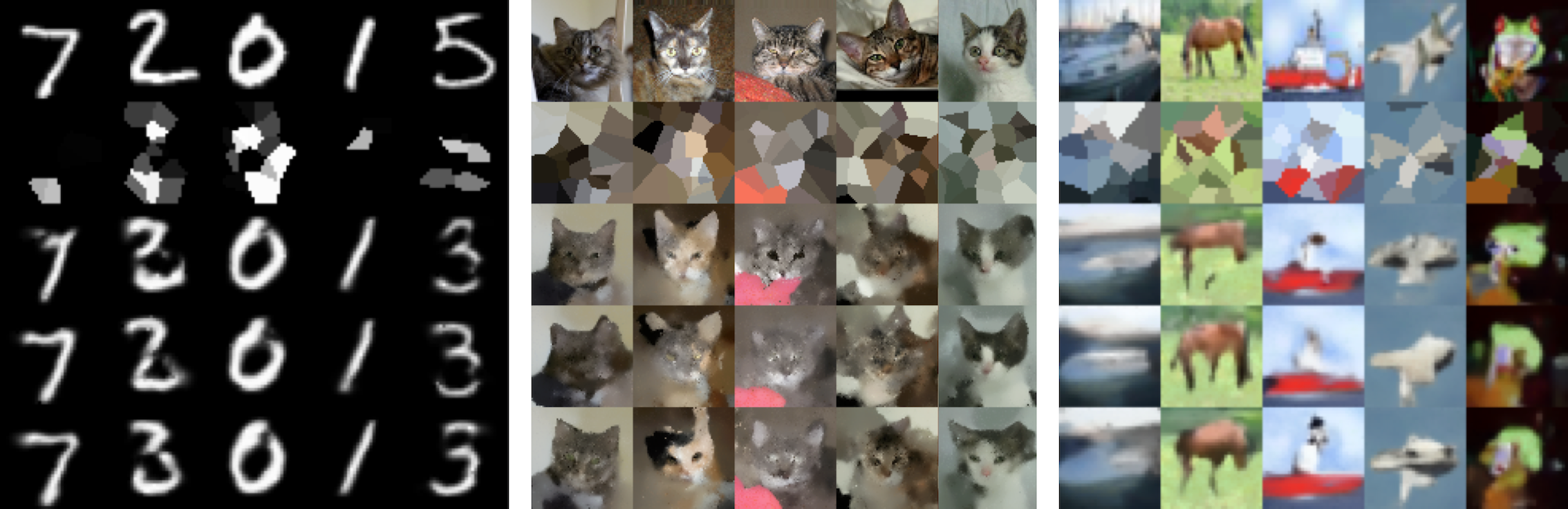}
    \vspace{1mm}
    \caption{\textbf{Image Samples.} Sample images generated by our learned model on three datasets (left: MNIST, middle: Cat Faces, right: CIFAR10) given $|S|=32$ observed pixels. Top row: ground-truth image from which $S$ is drawn. Row 2: A  nearest neighbor visualization of $S$ -- for each image pixel we assign it the color of the closest observed sample in $S$. Rows 3-5: Randomly sampled images from $p(I|S)$.}
    \figlabel{img_samples}
    \vspace{-3mm}
\end{figure*}

\noindent \textbf{Training Setup.} We vary the number of observed pixels $S$ randomly between 4 and 2048 (with uniform sampling in log-scale), while the number of query samples $Q$ is set to 2048. During training, the locations $\pos$ are treated as varying over a continuous domain, using bilinear sampling to obtain the corresponding value -- this helps our implementation be agnostic to the image resolution in the dataset. While we train a separate network $f_{\theta}$ for each dataset, we use the exact \emph{same} model, hyper-parameters \etc across them. 

\begin{figure}
    \centering
    \includegraphics[width=.95\linewidth]{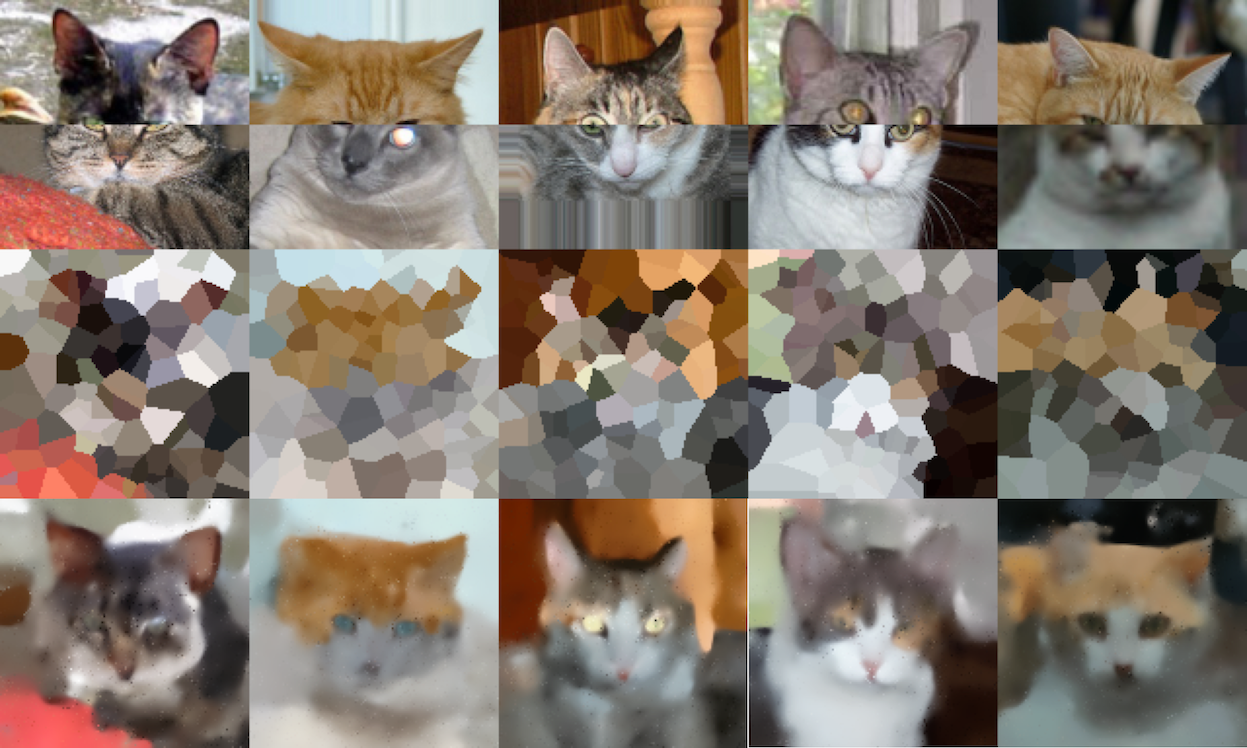}
    \vspace{1mm}
    \caption{\textbf{Image Composition.} Generation results when drawing pixels from two different images. Top row: the composed image from which $S$ is drawn. Row 2: A  nearest neighbor visualization of $S$. Row 3: Randomly sampled image from $p(I|S)$.}
    \figlabel{composition}
    \vspace{-3mm}
\end{figure}

\noindent \textbf{Qualitative Results: Mean Image Prediction.}
We first examine the expected image $\bar{I}$ inferred by our model given some samples $S$. We visualize in \figref{mean_cat} our predictions on the Cat Faces dataset using varying number of input samples. We observe that even when using as few as 4 pixels in $S$, our model predicts a cat-like mean image that, with some exceptions, captures the coarse color accurately. A very small number of pixels, however, is not sufficiently informative of the pose/shape of the head, which become more accurate given around 100 samples. As expected, the mean image becomes closer to the true image given additional samples, with the later ones even matching finer details \eg eye color, indicating that the distribution $p(I|S)$ reduces in variance as $|S|$ increases.

\noindent \textbf{Qualitative Results: Sampling Images.}
While examining the mean image assures us that our average prediction is meaningful, it does not inform us about samples drawn from $p(I|S)$. In \figref{img_samples}, we show  results on images from each of the three datasets considered using $|S|$=32 randomly observed pixel values in each case. We see that the sampled images vary meaningfully (\eg face textures) while preserving the coarse structure, though we do observe some artefacts \eg missing horse legs.

As an additional application, we can generate images by mixing pixel samples from different images. We showcase some results in \figref{composition} where we show one generated image given some pixels from top/bottom of two different images. We see that, despite some mismatch in the alignment/texture of the underlying faces, our model is able to compose them to generate a plausible new image.
\begin{figure}
    \centering
    \begin{minipage}{0.49\linewidth}
        \centering
        \includegraphics[width=0.99\textwidth]{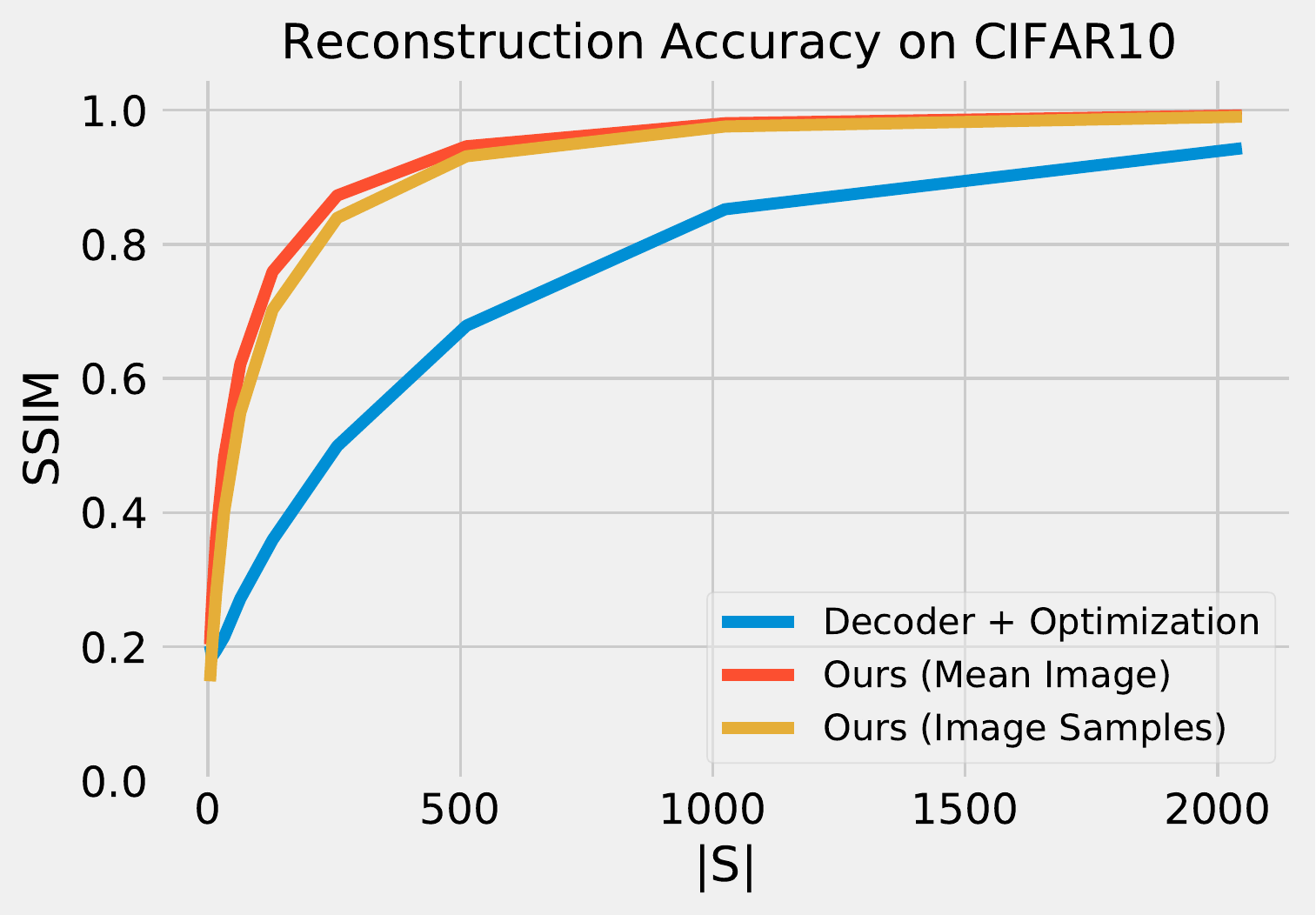} 
        \vspace{-3mm}
        \caption{Reconstruction Accuracy of generated images.}
        \figlabel{cifarrec}
    \end{minipage}\hfill
    \begin{minipage}{0.49\linewidth}
        \centering
        \includegraphics[width=0.99\textwidth]{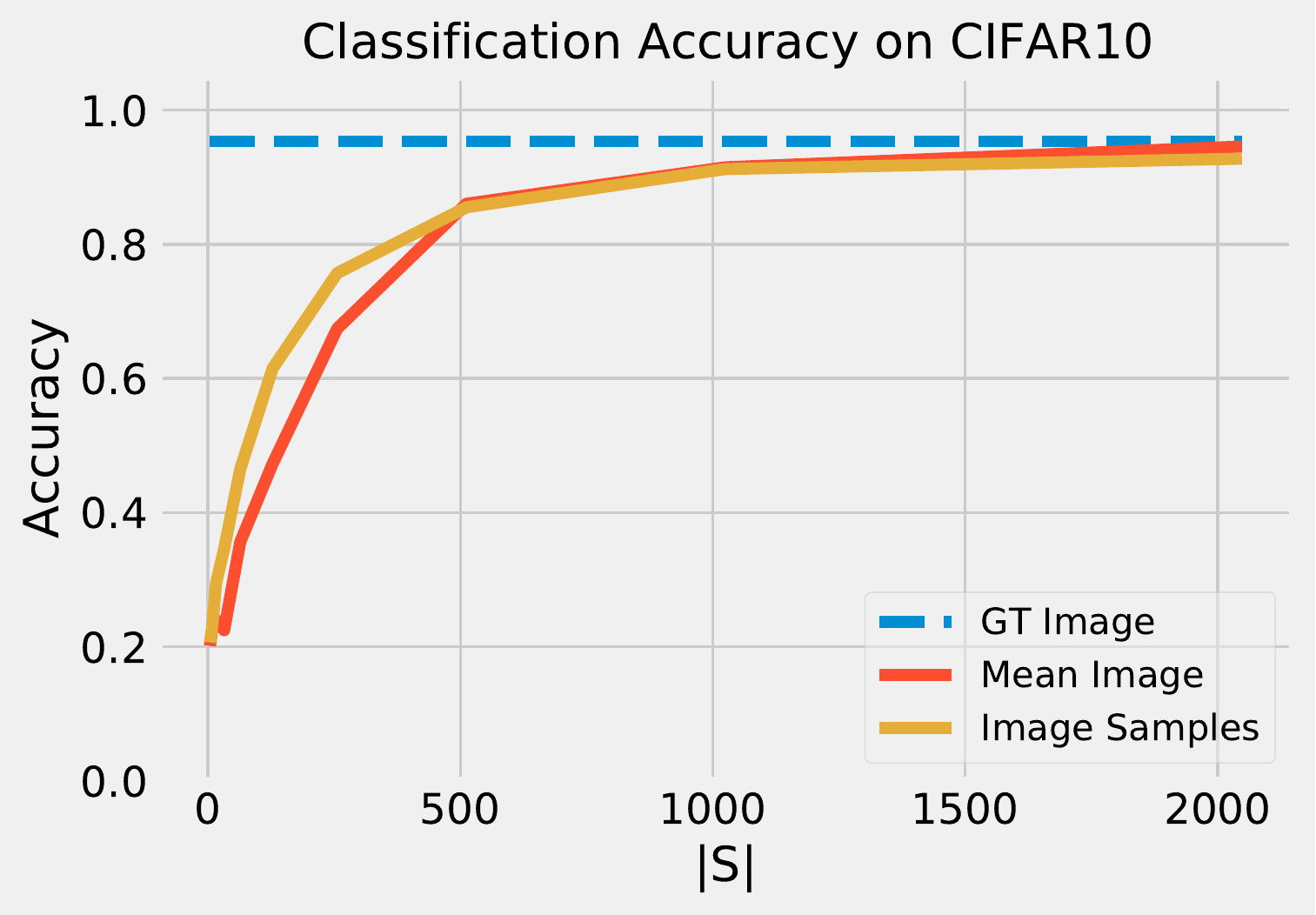} 
        \vspace{-3mm}
        \caption{Classification Accuracy of generated images.}
        \figlabel{cifarcls}
    \end{minipage}
    \vspace{-2mm}
\end{figure}

\begin{figure*}[t]
    \centering
    \includegraphics[width=.95\linewidth]{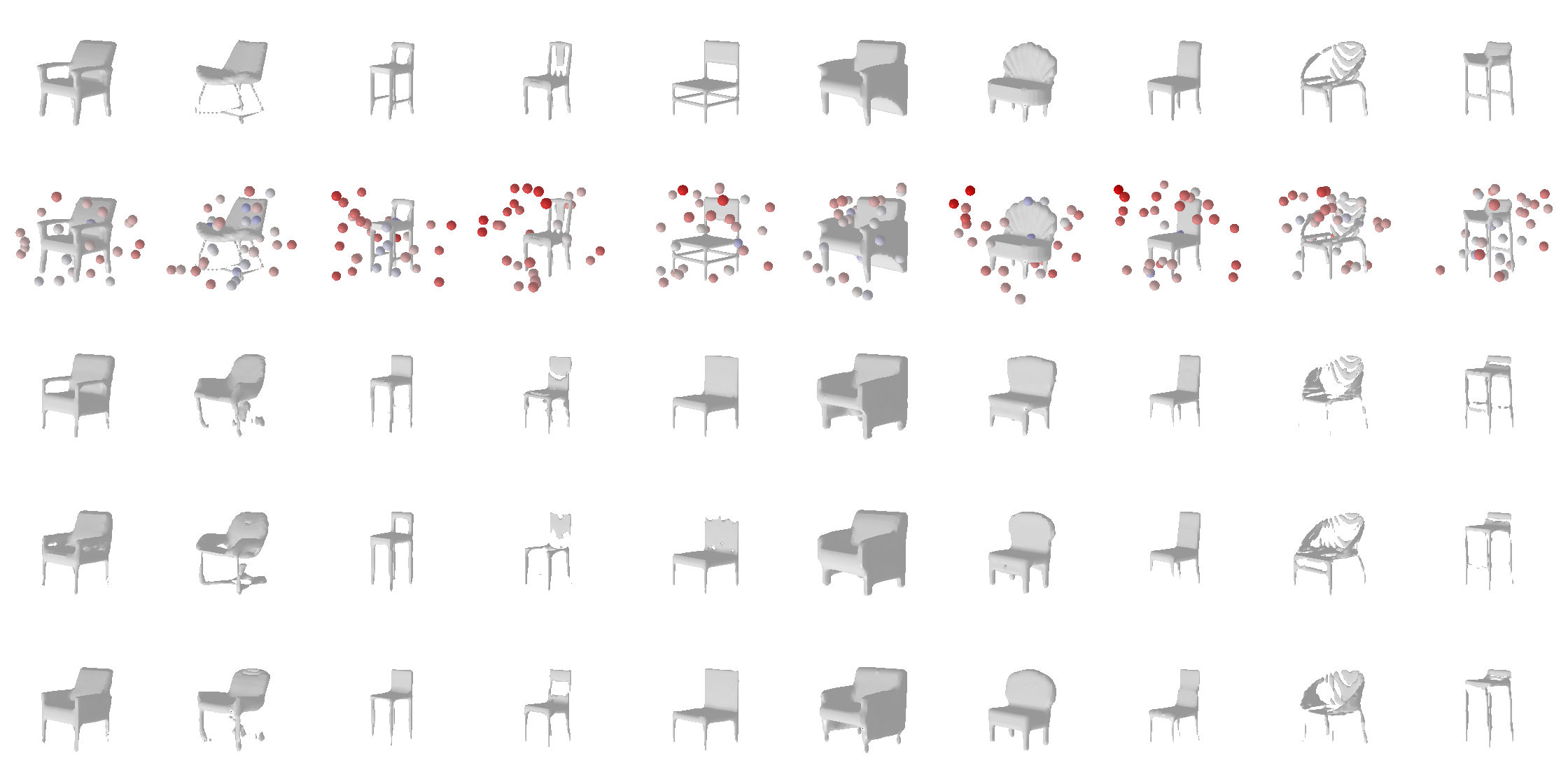}
    \vspace{1mm}
    \caption{\textbf{Shape Generation.} Sample 3D shapes generated given $|S|=32$ observed SDF values at random locations. Top row: ground-truth 3D shape. Row 2: A visualization of $S$ -- a sphere is centred at each position with color indicating value (red implies higher SDF). Rows 3-5: Randomly sampled 3D shapes from our predicted conditional distribution.}
    \figlabel{shape_pred}
\end{figure*}

\noindent \textbf{Reconstruction and Classification Accuracy.}
In addition to visually inspecting the mean and sampled images, we also quantitatively evaluate them using reconstruction and classification based metrics on the CIFAR10 dataset. 
First, we measure how similar our obtained images are to the underlying ground-truth image. \figref{cifarrec} plots this accuracy for varying size of $S$ -- we compute this plot using 128 test images, varying $|S|$ from 4 to 2048 for each. When reporting the accuracy for sampled images, we draw 3 samples per instance and use the average performance. We also report a baseline that uses a pretrained decoder(from a VAE) and optimizes the latent variable to best match the pixels in $S$ (see appendix for details). We observe that our predicted images, more so than the baseline, match the true image. Additionally, the mean image is slightly more `accurate' in terms of reconstruction than the sampled ones -- perhaps because the diversity of samples makes them more different.

We also plot the classification accuracy of the generated images in \figref{cifarcls}. To do so, we use a pretrained ResNet-18~\cite{resnet} based classifier and measure whether the correct class label is inferred from our generated images. Interestingly, we see that even if using images generated from as few as 16 pixels, we obtain about a 30\% classification accuracy (or over 60\% with 128 pixels). As we observe more pixels, the accuracy matches that of using the ground-truth images. Finally, we see that using the sampled images yields better results compared to the mean image,  as the sampled ones look more `real'.

%% file: nd.tex
\begin{figure}[t]
    \centering
    \includegraphics[width=.95\linewidth]{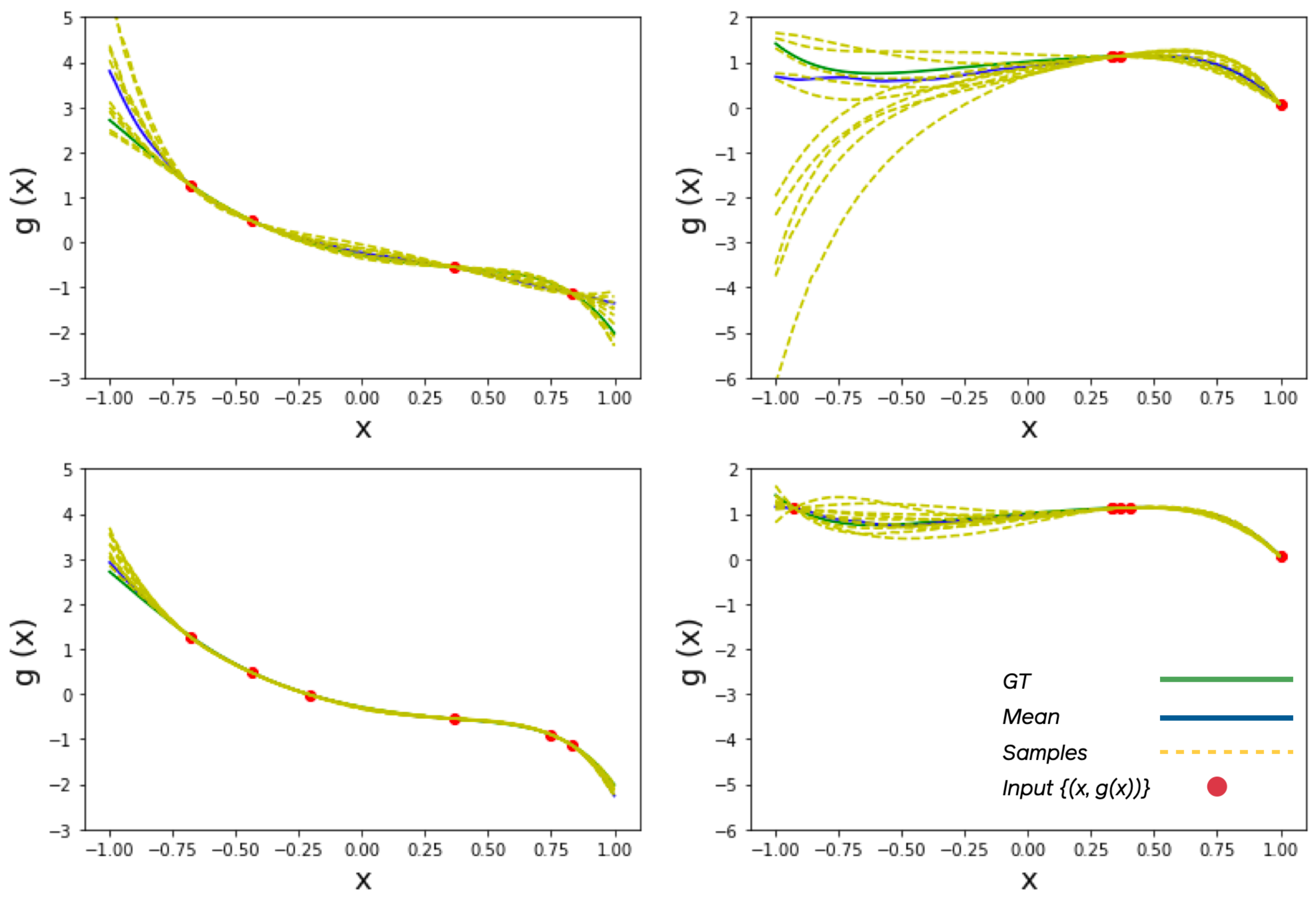}
    \caption{\textbf{Polynomial Prediction.} Mean and sampled polynomials generated by our learned model. Row 1: Predictions using $|S|=4$ samples (red dots). Row 1: Predictions using $|S|=6$.}
    \figlabel{poly_pred}
    \vspace{-1mm}
\end{figure}

\begin{figure}[t]
    \centering
    \includegraphics[width=.95\linewidth]{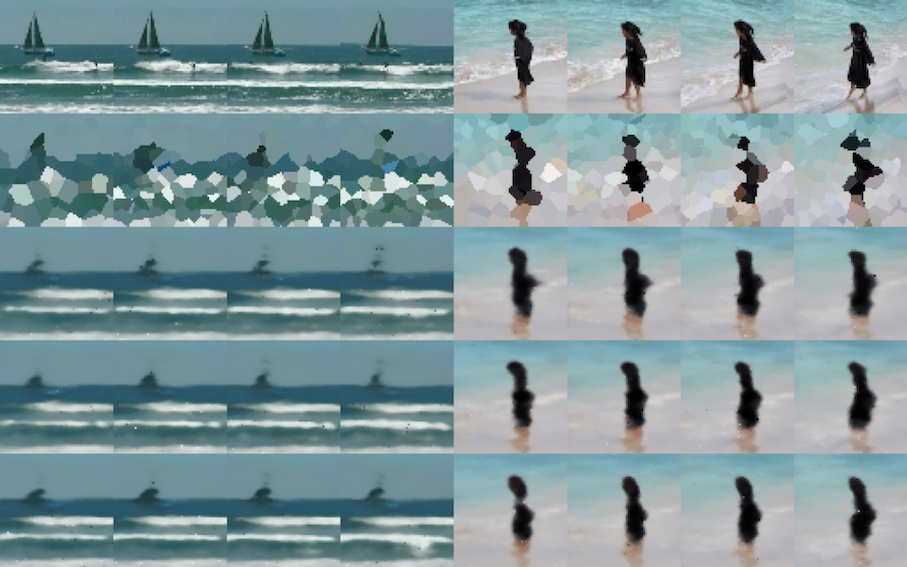}
    \caption{\textbf{Video Synthesis.} Sample videos generated by our model given $|S|$=1024 observed pixels across 34 frames. Top row: 4 uniformly sampled frames of the ground-truth video. Row 2: A nearest neighbor visualization of $S$. Rows 3-5: Randomly sampled videos from the predicted conditional distribution.}
    \figlabel{video_pred}
    \vspace{-1mm}
\end{figure}

\vspace{-1mm}
\section{Beyond Images: 1D and 3D Signals}
\vspace{-1mm}
\seclabel{nd}
While we leveraged our proposed framework for generating images given some pixel observations, our formulation is applicable beyond images. In particular, assuming the availability of (unlabeled) examples, our approach can learn to generate any dense spatial signal given some (position, value) samples. In this section, we empirically demonstrate this by learning to generate 1D (polynomial) and 3D (shapes and videos) signals using our framework. 

We would like to emphasize that across these settings, where we are learning to generate rather different spatial signals, we use the \emph{same} training objective and model design. That is, except for the dimensionality of input/output layers and distribution parametrization to handle the corresponding inputs/outputs $\pos \in \R^x, \val \in \R^v$, our model or learning objective is not modified in any way specific to the domain.

\vspace{-1mm}
\subsection{Polynomial Prediction}
\vspace{-1mm}
As an illustrative example to study our method, we consider a classical task -- given a sparse set of $(x, g(x))$ pairs, where $x, g(x) \in \R^1$, we want to predict the value of $g$ over its domain. We randomly generate $6$-degree polynomials, draw from 4 to 20 samples to obtain $S$, and learn $f_{\theta}$ to predict distribution of values at $|Q|$=20 query locations. One simplification compared to the model used for images is we use $B=1$ instead of $B=256$ (\ie a simple gaussian distribution) to parametrize the output distribution.

We visualize our predictions in \figref{poly_pred}, where the columns correspond to different polynomials, and the rows depict our results with varying number of inputs in $S$. We see that the various sample functions we predict are diverse and meaningful, while being constrained by the observed position, value pairs. Additionally, as the number of observations in $S$ increase, the variance of the function distribution reduces and matches the true signal more closely.

\vspace{-1mm}
\subsection{Generating 3D Shapes}
\vspace{-1mm}
We next address the task of generating 3D shapes represented  as signed distance fields (SDFs). We consider the category of chairs using models from 3D Warehouse~\cite{3DW}, leveraging the subset recommended by Chang \etal~\cite{chang2015shapenet}. We use the train/test splits provided by ~\cite{disn}, with 5268 shapes used for training, and 1311 for testing. We extract a SDF representation for each shape as a grid of size $64^3$, with each location recording a continuous signed distance value -- this dense representation is better suited for our approach compared to sparse occupancies.  Our training procedure is exactly the same as the one used for 2D images -- we sample the SDF grid at random locations to generate $S, Q$, with the number of samples in $S$ varying from 4 to 2048, and $|Q|$ being 2048. 

We present some \emph{randomly chosen} 3D shapes generated by our model when using $|S| = 32$ in \figref{shape_pred}.  While we actually generate a per-location signed distance value, we extract a mesh using marching cubes to visualize this prediction. As the results indicate, even when using only 32 samples from such a high-dimensional 3D spatial signal, our model is able to generate diverse and plausible 3D shapes. In particular, even though this is not explicitly enforced, our model generates symmetric shapes and the variations are semantically meaningful as well as globally coherent \eg slope of chair back, handles with or without holes. However, as our model generates the SDF representation, and does not directly produce a mesh, we often see some artefacts in the resulting mesh \eg disconnected components, which can occur when thresholding a slightly inconsistent SDF.

\vspace{-1mm}
\subsection{Synthesizing Videos}
\vspace{-1mm}
Lastly, we examine the domain of `higher-dimensional' images (\eg videos). In particular, we use the subset of `beach' videos in the TinyVideos dataset \cite{vondrick2016generating,thomee2016yfcc100m} (with a random $80 \% -20 \% $ train-test split) and train our model to generate video clips with 34 frames.
Note that these naturally correspond to 3D spatial signals, as the position $\pos$ includes timeframe $\in \R^1$ in addition to a pixel coordinate.

We train our model $f_{\theta}$ to generate the underlying signal distribution given sparse pixel samples where we randomly choose a frame and pixel coordinate for each sample. We empirically observe that due to the high complexity of the output space, using only a small number of samples does not provide significant information for learning generation. We therefore train our model using more samples than the image generation task -- varying $|S|$ between 512 to 2048 (this  corresponds to 30 pixels per frame).

We present representative results in \figref{video_pred} but also encourage the reader to see the videos in the project page. Our model generates plausible videos with some variation \eg flow of waves and captures the coarse structure of the output well. However, the predictions lack precise detail. We attribute this to the limited number of pixels we can generate autoregressively (see discussion in \secref{sampling} on memory bottlenecks) and hypothesize that a higher number maybe needed for modeling these richer signals.

%% file: discussion.tex
\vspace{-1mm}
\section{Discussion}
\vspace{-1mm}
We proposed a probabilistic generative model capable of generating images conditioned on a set of random observed pixels, or more generally, synthesizing spatial signals given sparse samples. At the core of our approach is a learned function that predicts value distributions at any query location given an arbitrary set of observed samples. While we obtain encouraging results across some domains, there are several aspects which could be improved \eg scalability, perceptual quality, and handling sparse signals. To allow better scaling, it could be possible to generalize the outputs from distributions over individual pixels to those over a vocabulary of tokens encoding local patches or investigate strategies to better select conditioning subsets (\eg nearest samples). The perceptual quality of our results could be further improved and incorporating adversarial objectives maybe a promising direction.  Finally, while our framework allowed generating  pixel values, we envision that a similar approach could  predict other dense properties of interest \eg semantic labels, depth, generic features.

%% file: appendix.tex
\section*{Appendix}

\paragraph{Log-likelihood under Value Distribution.}
The predicted value distribution for a query position $\pos$ is of the form $p(\val;\omega)$, where $\omega \equiv \{(q^b, \mu^b, \sigma^b)\}_{b=1}^B$. We reiterate $q^b \in \R^1$ is the probability of assignment to bin $b$, $c^b + \mu^b$ is the mean of the corresponding gaussian distribution with uniform variance $\sigma^b \in \R^1$.

Under this parametrization, we compute the log-likelihood of a value $\val^{*}$ by finding the closest bin $b^{*}$, and computing the log-likelihood of assignment to this bin as well as the log-probability of the value under the corresponding gaussian. We additionally use a weight $\alpha = 0.1$ to balance the classification and gaussian log-likelihood terms. 
\begin{gather*}
    b^* = \text{argmin}_b~\|v^* - c^b\| \\
    \log p(v^*; \omega) \equiv \log q^{b^{*}} - \alpha (\log \sigma^{b^{*}} + (\frac{\val^* - c^{b^{*}} - \mu^{b^{*}}}{\sigma^{b^{*}}})^2)
\end{gather*}

\paragraph{VAE Training and Inference.}
We train a variational auto-encoder~\cite{vae} on the CIFAR10 dataset with a bottleneck layer of dimension $4 \times 4 \times 64$ \ie spatial size $4$ and feature size $64$. We consequently obtain a decoder $\mathcal{D}$ which we use for inference given some observed samples $S$. Specifically, we optimize for an optimal latent variable the minimizes the reconstruction loss for the observed samples (with an additional prior biasing towards the zero vector). Denoting by $I(\pos)$ the value of image $I$ (bilinearly sampled) at position $\pos$, the image $I^*$ inferred using a decoder $D$ by optimizing over $S$ can be computed as: 
\begin{gather*}
    z^* = \text{argmin}_z~L(D(z), S) + 0.001*\|z\|^2; ~~ I^* = D(z^*) \\
    L(I, \{(\pos_k, \val_k)\}) = \mathop{\mathbb{E}}_{k} \|I(\pos_k) - \val_k \|_1
\end{gather*}